\documentclass{article}

\PassOptionsToPackage{numbers}{natbib}

\usepackage[final]{neurips_data_2023}

\newcommand{\reviewscount}{325}
\newcommand{\datasetname}{\textsc{CSMeD}}
\newcommand{\ftdatasetname}{\textsc{\datasetname-ft}}

\usepackage[utf8]{inputenc} %
\usepackage[T1]{fontenc}    %
\usepackage{hyperref}       %
\usepackage{url}            %
\usepackage{booktabs}       %
\usepackage{amsfonts}       %
\usepackage{nicefrac}       %
\usepackage{microtype}      %
\usepackage{xcolor}         %
\usepackage{graphicx}
\usepackage{enumitem}
\usepackage{listings}       %

\usepackage{amsmath,amsthm}
\theoremstyle{definition}
\newtheorem{definition}{Definition}[section]

\usepackage{multirow}
\usepackage{scrextend}

\definecolor{verbgray}{gray}{0.9}
\lstnewenvironment{code}{%
  \lstset{backgroundcolor=\color{verbgray},
  frame=single,
  framesep=5pt,
  framerule=0.5pt,
  basicstyle=\ttfamily,
  columns=fullflexible}}{}
\definecolor{shadecolor}{rgb}{.9, .9, .9}

\title{\datasetname: Bridging the Dataset Gap in Automated Citation Screening for Systematic Literature Reviews}

\author{
\hspace{-.4cm}Wojciech Kusa$^{1}\thanks{Corresponding author: \texttt{wojciech.kusa@tuwien.ac.at}}$ {\normalfont, } \enskip Oscar E. Mendoza$^{2}${\normalfont, } \enskip
\textbf{Matthias Samwald}$^{3}${\normalfont, } \enskip \textbf{Petr Knoth}$^{4}${\normalfont, } \enskip \textbf{Allan Hanbury}$^{1}$ \\
\hspace{-.4cm}$^{1}$TU Wien \enskip $^{2}$University Milano-Bicocca \enskip $^{3}$Medical University of Vienna \enskip $^{4}$The Open University  \\
}

\begin{document}

\maketitle

\begin{abstract}

Systematic literature reviews (SLRs) play an essential role in summarising, synthesising and validating scientific evidence.
In recent years, there has been a growing interest in using machine learning techniques to automate the identification of relevant studies for SLRs. 
However, the lack of standardised evaluation datasets makes comparing the performance of such automated literature screening systems difficult.
In this paper, we analyse the citation screening evaluation datasets, revealing that many of the available datasets are either too small, suffer from data leakage or have limited applicability to systems treating automated literature screening as a classification task, as opposed to, for example, a retrieval or question-answering task.  
To address these challenges, we introduce \datasetname, a meta-dataset consolidating nine publicly released collections, providing unified access to \reviewscount~SLRs from the fields of medicine and computer science. 
\datasetname~serves as a comprehensive resource for training and evaluating the performance of automated citation screening models.
Additionally, we introduce \ftdatasetname, a new dataset designed explicitly for evaluating the full text publication screening task.
To demonstrate the utility of \datasetname, we conduct experiments and establish baselines on new datasets.
\end{abstract}

\section{Introduction}

\emph{Systematic literature reviews} (\emph{SLRs}, or meta-reviews) are a critical tool in scientific research, used for synthesising and summarising evidence from multiple studies. 
The SLR process involves several stages, including \emph{citation screening} (\emph{CS}, or selection of primary studies) which is, in itself, a time-consuming step~\cite{nussbaumer2021resource,clark2020full}.
CS involves identifying studies relevant to the SLR based on a set of, often complex, inclusion and exclusion criteria (e.g., the study must be examining the efficacy of Drug X on Condition Y).

In recent years, there has been an increasing interest in automating the SLR process~\cite{VanDinter2021,OMara-Eves2015,Norman2020,hannousse2022semi,Alharbi2019}, with works often focusing on improving the CS step by (a) using machine learning (ML)~\cite{lee2023pgb,lee2023sr}, (b) natural language processing (NLP)~\cite{Howard2016a,Kontonatsios2017AScreening,VanDinter2021c}, and (c) information retrieval (IR)~\cite{scells2018generating,wang2023can} techniques. 
Automated CS systems have the potential to significantly reduce the time and resources required for this critical step, thereby speeding up the SLRs production~\cite{tsafnat2013automation}.

The development of standards provides invaluable resources for evaluating and comparing different models. 
Benchmarks, such as BEIR~\cite{Thakur2021BEIR:Models}, GLUE~\cite{wang2018glue} or BLURB~\cite{DBLP:journals/health/GuTCLULNGP22} have shown improvements in reproducibility and progress tracking of machine learning models in various domains. 
Unfortunately, in the context of SLR automation, the absence of standard benchmarks and evaluation methodologies still hampers progress and inhibits the development of reliable and effective solutions.

With the fast-evolving landscape of machine learning, identifying state-of-the-art performance has become especially challenging and inefficient in the context of CS.
The notorious proliferation of small custom CS datasets and single-usage evaluation approaches further exacerbates this issue. 
We show that current CS datasets exhibit several shortcomings that hinder their applicability for comprehensive and standardised evaluations.
These datasets are poorly documented, with most lacking datasheets, clear licenses and terms of use. 
In addition, the limited applicability of older datasets arises from their small size and lack of crucial metadata, restricting their use to classification tasks. 
Finally, data leakage and dataset overlap is another issue, with some SLRs present in multiple collections.

To address these limitations, we present \datasetname~(\textbf{C}itation \textbf{S}creening \textbf{Me}ta-\textbf{D}ataset), a comprehensive collection of CS datasets that can be used to benchmark and evaluate automated screening systems. %
Our collection builds upon nine existing datasets, and a new dataset for evaluating the full text classification task, counting \reviewscount~SLRs from the fields of medicine and computer science. 
Thanks to the data harmonisation, our new collection can mitigate the issues of lack of canonical splits, limited applicability, and dataset overlap.
Our contributions are as follows:

\begin{enumerate}
  \item  We create \datasetname, a meta-collection of nine datasets comprising \reviewscount~SLRs.
    \datasetname~is built upon \textsc{BigBio}~\citep{fries2022bigbio} and can be used to evaluate and benchmark automated CS systems.
    We also provide a comprehensive summary of existing citation screening datasets. 

    \item We extend \datasetname~with additional metadata after analysing issues on the existing collections and previous evaluation frameworks. %
    Our extended dataset can be used to evaluate CS as question answering or textual pairs classification tasks.
    
    \item  Using new metadata, we introduce \ftdatasetname, a new dataset for the task of full text screening. 
    To the best of our knowledge, this is the first dataset designed explicitly on screening long documents in SLR.
    This dataset can be used for the evaluation of the inference capabilities based on a very long context (4,000+ words).
    
\end{enumerate}

The remainder of the paper is structured as follows. 
In Section~\ref{sec:task_formulation}, we define the task of citation screening for systematic literature reviews.
Section~\ref{sec:related_work} provides an overview of related work in SLR automation and available benchmarks. 
Section~\ref{sec:meta_dataset_collection} describes the \datasetname~meta-dataset in detail, including its creation, analysis and extension. 
In Section~\ref{sec:full_text_dataset}, we introduce the full text screening dataset together with baseline results on this dataset, and in Section~\ref{sec:discussion}, we discuss the implications of our work and potential extensions.

\section{Task formulation} \label{sec:task_formulation}

We start by introducing the task of citation screening for SLRs and presenting the notation used for its formulation. 
An SLR is characterised by various attributes, including the title, abstract, research question $\mathcal{RQ}$, and eligibility criteria $\mathcal{C}$. 
We refer to all these attributes as the SLR protocol.
Eligibility criteria comprise a set of rules and conditions that a document must meet for inclusion in the SLR.
Given a large pool of documents denoted as $\mathcal{D}$, the main goal of automated citation screening is to assist researchers in identifying relevant publications for inclusion in an SLR.
Each document $d \in \mathcal{D}$ has attributes such as its title, abstract, main content, authors, and publication year. 
The task of CS for SLRs can be formally defined as follows:

\begin{definition}[CS]
Given a set of documents $\mathcal{D}$ and a set of eligibility criteria $\mathcal{C}$, the task of CS for SLR is to determine for each document $d \in \mathcal{D}$ whether it satisfies the criteria $\mathcal{C}$.
This decision can be represented as a binary label $y_d \in \{0, 1\}$, where $y_d = 1$ if document $d$ satisfies the criteria $\mathcal{C}$, and $y_d = 0$ otherwise.
\end{definition}

\begin{figure}[t]
    \centering
    \includegraphics[width=\textwidth]{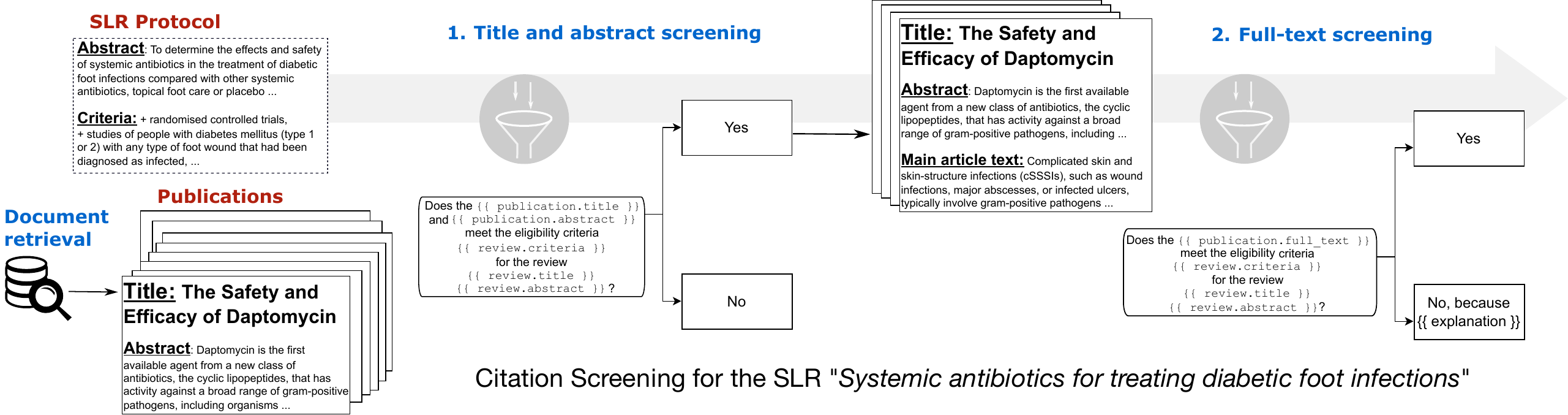}
    \caption{Illustration of the citation screening process, separated into two tasks (1) title and abstract screening and (2) full text screening. Tasks are represented as a specific example of question-answering when a single question asks for a fullfilment of all eligibility criteria $\mathcal{C}$ at once. 
    }
    \label{fig:screening}
\end{figure}

It is important to note that the manual CS is conducted in two steps, as shown in Figure~\ref{fig:screening}: title and abstract screening and full text screening. 
In the first step, the relevance of each document is evaluated based on its title and abstract, while in the second step, a more thorough assessment is performed by examining the full text of the document.

\paragraph{Document retrieval}
The initial step involves document retrieval, which aims to generate a set of potentially relevant documents $\mathcal{D'} \subseteq \mathcal{D}$ given $\mathcal{RQ}$.
This step commonly involves querying bibliographic databases with specific keywords and Boolean expressions.
We can formulate this step as a retrieval function $r$, such that $r(\mathcal{RQ}, \mathcal{C}) = \mathcal{D'}$.
However, the retrieved set $\mathcal{D'}$ may contain a large number of false positives (irrelevant documents).

\paragraph{Binary classification for relevance prediction}

Following document retrieval, the primary task is to assess the relevance of each document in the set $\mathcal{D'}$ concerning the eligibility criteria $\mathcal{C}$. 
This is conducted in two stages, differing in which attributes of documents are considered (titles and abstracts \textit{vs.} full texts).
We treat this as a binary classification problem, where each document $d \in \mathcal{D'}$ is assigned a binary label $y_d \in \{0, 1\}$ to indicate its relevance ($y_d = 1$) or irrelevance ($y_d = 0$) to the SLR per the criteria $\mathcal{C}$.

\paragraph{Question answering for relevance}

An alternative formulation of the citation screening task is to frame it as a question-answering problem.
In this approach, we transform the eligibility criteria $\mathcal{C}$ into a set of questions $\mathcal{Q}=\{q_1, \cdots, q_{\vert C \vert}\}$, where each question $q_k$ corresponds to a specific criterion in $\mathcal{C}$.
For each document $d \in \mathcal{D'}$, we obtain a set of predicted answers $\hat{A}^d = \{\hat{a}_{k}^d \vert \text{meets}(q_k,\hat{a}_k^d) \}$, where $\text{meets}(q_k,\hat{a}_k^d)$ denotes that the document $d$ should meet the criterion expressed by the question $q_k$.
The final relevance label $\hat{y}_d$ of a document $d$ can be determined by aggregating the predicted answers $\hat{A}^d$ using a logical combination function, such as the logical AND operation.

This question-answering formulation offers a more fine-grained assessment of a document's relevance concerning various aspects of the eligibility criteria $\mathcal{C}$. 
Other similar formulations of the CS task include document ranking or natural language inference (NLI).

\section{Related work} \label{sec:related_work}

We first motivate the work by providing context on the importance of SLRs and then focus on reviewing citation screening automation methods.
Finally, we outline limitations of existing CS datasets.

\subsection{Systematic literature reviews}

SLRs are particularly important in the medical domain~\cite{Jo2009}.
The Cochrane Collaboration,\footnote{\url{https://www.cochrane.org}} the largest organisation responsible for creating SLRs in medicine, has created the foundations of Evidence-Based Medicine~\cite{higgins2019cochrane}.
There are more than 220,000 records published between 2000 and 2022 tagged as SLRs in PubMed\footnote{\url{https://pubmed.ncbi.nlm.nih.gov/}} meaning that, on average, there were 10,000 SLRs published per year. 

As SLRs focus on reproducibility and finding all relevant evidence about a given topic, the traditional framework involves tasks mainly done manually. 
It includes steps like defining the search strategy (designing complex Boolean queries) or the screening of every document by at least two reviewers, resulting in an average production time of more than one year~\cite{Tsafnat2018}.

Previous research focused on evaluating automation capabilities for several steps of the traditional framework, such as citation screening (CS)~\cite{Cohen2006,Howard2016a,tsafnat2013automation}, search query (re-)formulation~\cite{scells2018generating,scells2021comparison}, data extraction~\cite{nye2018corpus}, SLR summarisation~\citep{wang-etal-2022-overview} or generation of reviews based on the title~\citep{yun2023appraising}.

\subsection{Citation screening automation}

As described in Section~\ref{sec:task_formulation}, CS can be seen as a binary classification problem. 
However, due to a large number of retrieved studies, the significant class imbalance, and the need to identify \emph{nearly all} relevant documents, this task is inherently complex.
Screening automation is a general term for various approaches aimed at reducing workload during the CS stage~\cite{OMara-Eves2015}.
These approaches can be classified as either screening reduction, which involves using classification or ranking algorithms to automatically exclude non-relevant publications or screening prioritisation, which focuses on ranking relevant records earlier in the screening process \cite{Norman2020}. 
Automated screening systems leverage techniques from NLP, ML, IR, and statistics, all with the common objective of reducing manual screening time.
The disparity in strategies from different fields hinders direct comparison and benchmarking. 
Next, we discuss some points of disagreement.

NLP approaches typically focus on the level of individual SLRs, treating each review as an independent dataset; whereas IR approaches would consider a set of reviews as a collection, the topics of the reviews analogous to queries, and report aggregated evaluation. 
Moreover, different publications across various venues adopt diverse evaluation measures, making even more complex the assessment of similar, if not identical, tasks.
Evaluation of automatic approaches traditionally relies on binary relevance ratings, very often obtained from the title and abstract screening~\cite{OMara-Eves2015,Kanoulas2019CLEF2T}. 
When the screening problem is treated as a ranking task, such as screening prioritisation or stopping prediction; the performance is measured in terms of rank-based metrics and metrics at a fixed cut-off, such as $nDCG@n$, $Precision@n$, and \emph{last relevant found}~\cite{Scells2017,Howard2020}. 
On the other hand, when the screening problem is treated as a classification task, the performance in this case is measured based on the confusion matrix and the notions of Precision and Recall are commonly used \cite{kusa2023vombat,OMara-Eves2015}.
One challenge arising from these two distinct approaches is the difficulty in going beyond simple effectiveness measures and comparing the real-world savings for users. 
Further details on datasets and evaluation approaches can be found in a comprehensive review in the Appendix \ref{app:datasets}.

\subsection{Limitations of existing datasets}

Through our review (see Appendix~\ref{app:datasets}), we identified twelve CS datasets reported in former research papers, of which ten have been publicly released.
During this analysis, we identified several shortcomings; some are also prevalent in other machine learning problems. 
Below, we summarise our findings, highlighting the key issues. 

\paragraph{Poor documentation}
One major concern with previous datasets is the lack of sufficient documentation.
None of the datasets we examined implement a datasheet~\cite{DBLP:journals/cacm/GebruMVVWDC21}, which is an essential tool for ensuring transparency and reproducibility.
Additionally, seven datasets do not provide clear licenses or terms of use. 
An inconsistency was also found for one of the datasets~\cite{Scells2017} in terms of the number of the available content: the paper states 93 SLRs, but we found a list of 176 reviews on the corresponding GitHub repository. 

\paragraph{Limited applicability}
Previous datasets are often small and lack crucial metadata like SLR research question or eligibility criteria, limiting their use to only evaluation of classification tasks. 
Older datasets typically provide only the title of the review, which limits their applicability for the comprehensive evaluation of neural language understanding models.
The most widely used dataset to date~\cite{Cohen2006} was released in 2006. 
As ML and NLP techniques continue to advance rapidly, it is crucial to have up-to-date datasets that reflect the complexities and nuances of the current research landscape.

\paragraph{Lack of canonical splits}
Another significant challenge of previous datasets is the absence of canonical train-test splits. Depending on the field of research, practices may vary. As discussed before,
in the ML and NLP domains, the prevailing practice is to use inter-review splits, where each review is treated as an individual dataset, and a set of citations is selected for training and testing. 
Conversely, IR publications often report intra-review splits, treating each review as a ``topic'' or query, and averaging the results across multiple queries. 

In this sense,
only the three TAR\footnote{TAR stands for Technology-Assisted Reviews and was a shared task organised at CLEF between 2017 and 2019 by \citet{Kanoulas2017CLEFOverview,Kanoulas2018CLEFOverview,Kanoulas2019CLEF2T}.} datasets contain pre-defined canonical splits, yet only at the intra-review level. For three other datasets~\cite{Cohen2006,wallace2010semi,Howard2016a}, previous works have demonstrated significant variability in model evaluation based on the selection of cross-validation splits, particularly for the smallest datasets that contain a limited number of relevant documents~\cite{VanDinter2021c,Kusa2022AutomationStudy}.
The lack of standardised splits, especially in collections with fewer SLRs, makes it challenging to compare different approaches and hinders the fair evaluation of models' performance.

\paragraph{Dataset overlap}
We also evaluated overlapping throughout the previous datasets and discovered that at least 11 SLRs were present in multiple collections~\cite{Scells2017,Kanoulas2017CLEFOverview,Kanoulas2018CLEFOverview,Kanoulas2019CLEF2T}. 
Additionaly, the TAR 2019 dataset contains three SLRs that are present both in its training and test splits, accounting for approximately 6\% of the test partition~\cite{Kanoulas2019CLEF2T}. 
While this overlap is not a significant concern when evaluating unsupervised methods like BM25~\cite{robertson1995okapi}, it poses a potential threat to conducting fair comparisons with large language models (LLMs). 
Machine learning models, and especially LLMs, have the capability to memorise their training data, making it critical to address dataset overlap to ensure unbiased evaluations~\cite{lm-contamination} (see Appendix~\ref{app:dataset_overlap} for a detailed analysis of the overlapping in previous datasets).

\paragraph{Lack of common evaluation}
Another notable deficiency among the previous datasets is the absence of a common set of evaluation measures. 
Only the three TAR datasets provide scripts for evaluating submissions. 
For example, the most widely used dataset by~\citet{Cohen2006} was evaluated using several disparate evaluation measures such as $WSS$~\cite{Cohen2006}, $AUC$ or $Precision@r\%$. 
However, recent research has exposed limitations and problems with both $WSS$ and $AUC$ as metrics for this task~\cite{KUSA2023200193}.

\paragraph{Availability in biomedical benchmarks}
Recent efforts have focused on creating larger collections of more diverse datasets to evaluate the performance of biomedical NLP models. These efforts include benchmarks like BLUE~\cite{peng-etal-2019-transfer}, HunFlair~\cite{weber2021hunflair}, BLURB~\cite{DBLP:journals/health/GuTCLULNGP22}, and BigBio~\cite{fries2022bigbio}, which provide datasets and tasks for evaluating biomedical language understanding and reasoning. Additionally, there are biomedical datasets geared towards prompt-based learning and evaluation of few and zero-shot classification, such as Super-NaturalInstructions~\cite{wang-etal-2022-super} and  BoX~\cite{parmar-etal-2022-boxbart}.
Out of all benchmarks mentioned above, only BoX contains one CS dataset covering five SLRs, however, this dataset is private.
Coverage for other SLR tasks is also limited.

To summarise, previous datasets exhibit certain drawbacks that limit their suitability for comprehensive and standardised evaluation.
While the TAR 2017-19 collections stand out as the only ones containing canonical splits and a set of evaluation measures, some of their topics overlap with another previous dataset~\cite{Scells2017}, and we also identified data leakage in the newest TAR 2019 dataset.
Consequently, we believe that developing a new collection is necessary to address these issues and establish a robust foundation for evaluation of CS and SLR automation.

\section{The \datasetname~meta-dataset} \label{sec:meta_dataset_collection}

The recent advancements and paradigm shifts in NLP and ML; with the extensive use of pre-trained models and transfer learning~\cite{lewis-etal-2020-bart,dong2019unified}, and the more recent prompt-based learning~\cite{Liu2021Pre-trainProcessing,brown2020language}; have significantly transformed the field and enhanced the predictive capabilities of models across various tasks.
Inspired by the success of benchmark collections in the field of biomedical NLP, we conducted a thorough review of available datasets and benchmarks to identify the most representative datasets for the task of citation screening, finding that this task is heavily underrepresented. 
The available datasets still primarily cater to training supervised algorithms, lacking the scale and granularity necessary to evaluate state-of-the-art models.
To address these limitations and provide a more comprehensive resource for training and evaluating data-centric methods in SLR automation, we create \datasetname
, consolidating nine previously released collections of SLRs.
We further extend a subset of SLRs in \datasetname~with additional metadata coming from the review protocol.

Our data analysis methodology involved creating visualisations and summary tables based on curated datasets.
We analyse dataset statistics like available data splits, licensing information, dataset and reviews size as well as dataset overlap.
This allows us to provide both a detailed view of individual reviews and an overview of datasets containing multiple reviews (see Appendix~\ref{app:visualisations} for further details on visualisations).

\subsection{Dataset creation}
Currently, nine out of ten public CS datasets we identified have been included in \datasetname, with the remaining one to be included. 
We provide a summary of the datasets in Table \ref{tab:csmed_datasets}, and further details can be found in Appendix~\ref{app:datasets}.
In total, \datasetname~consists of \reviewscount~SLRs, making it the largest publicly available collection in this domain and the only one providing access to the datasets via a harmonised API.

\begin{table}[ht]
\centering
\caption{A list of source citation screening datasets included in the \datasetname. The first four datasets contain non-Cochrane SLRs, whereas the other five are based on Cochrane reviews. `Avg. ratio of included' column presents ratio of included publication from the title and abstract screening stage, `Avg. size' refers to averaged across SLRs document count in the dataset. The `Additional data' column describes if the review contains metadata other than coming from the citation list: (A): Search queries, (B): Review protocol containing review title, abstract and search strategy, (C): Review updates consisting of changes to included papers. `DTA' stands for diagnostic test accuracy reviews. $^{\ddag}$~--~indicates a discrepancy in the number of reviews in the paper versus the GitHub repository. $^{\dagger}$~--~indicates the total count of reviews from all nine datasets before duplicates were removed.}
\label{tab:csmed_datasets}
\resizebox{0.84\textwidth}{!}{%
\begin{tabular}{@{}cccrrcc@{}}
\toprule
\begin{tabular}[c]{@{}c@{}}Source\end{tabular} & 
\# reviews & Domain & 
\begin{tabular}[c]{@{}c@{}}Avg.\\ size\end{tabular} & 
\begin{tabular}[c]{@{}c@{}}Avg. ratio\\ of included\end{tabular} &
\begin{tabular}[c]{@{}c@{}}Additional\\data\end{tabular} & 
\begin{tabular}[c]{@{}c@{}}Cochrane\\reviews\end{tabular} \\ \midrule
\cite{Cohen2006} & 15 & Drug & 1,249 & 7.7\% & --- & ---  \\
\cite{wallace2010semi} & 3 & Clinical & 3,456 & 7.9\% & ---  & ---  \\
\cite{Howard2016a} & 5 & Mixed & 19,271 & 4.6\% & --- & ---  \\
\cite{hannousse2022semi} & 7 & Comp. Science & 340 & 11.7\% & B & ---  \\ \midrule
\cite{Scells2017} & 93/176$^{\ddag}$ & Clinical & 1,159 & 1.2\% & A &  $\checkmark$  \\
\cite{Kanoulas2017CLEFOverview} & 50 & DTA & 5,339 & 4.4\% & B &  $\checkmark$  \\
\cite{Kanoulas2018CLEFOverview} & 30 & DTA & 7,283 & 4.7\% & B &  $\checkmark$  \\
\cite{Kanoulas2019CLEF2T} & 49 & Mixed & 2,659 & 8.9\% & B &  $\checkmark$  \\
\cite{Alharbi2019} & 25 & Clinical & 4,402 & 0.4\% & C & $\checkmark$  \\ \midrule
Total & 360$^{\dagger}$  & & 3,471 & 4.4\% \\
\bottomrule
\end{tabular}%
}
\end{table}

To ensure interoperability and facilitate the ease of use, we designed data loaders for the datasets in accordance with the BigBio text classification schema~\cite{fries2022bigbio}. 
This choice offers several advantages.
BigBio has the largest coverage of biomedical datasets and supports access to the datasets via API.
Moreover, it is compatible with popular libraries such as Hugging Face's datasets~\cite{lhoest2021datasets} and the EleutherAI Language Model Evaluation Harness~\cite{eval-harness}, thereby reducing the experimental costs.%

Taking advantage of the lists of publications that most of the sources of datasets share as PubMed IDs, we extend the BigBio data loaders to enable the download of PubMed articles. 
Our harmonised data loaders selectively load the PubMed articles that are a part of each dataset.
The single exception is the dataset by \citet{hannousse2022semi}, which is the only publicly available collection of non-medical SLRs.
For this dataset, we extract the content using the SemanticScholar API.\footnote{\url{https://www.semanticscholar.org/product/api}}
As a result, \datasetname~serves also as the first resource that gathers SLRs from diverse domains.

\subsection{Extending metadata} \label{sec:prospective}
Our goal at this stage is not to create yet another gold standard dataset for SLRs, but rather improve the quality of current data and provide insights into promising avenues for future research. 
We begin by presenting the possibilities of extending the subset of Cochrane SLRs to experiment with screening beyond supervised classification.

We then categorise \datasetname~datasets into two groups: (1) datasets containing Cochrane medical SLRs and (2) datasets comprising other SLRs. 
This distinction is made because from following the Cochrane protocol, more extensive information on the review is provided. 
We use the additional data available on reviews websites to extend \datasetname.
Among the new information, we find the eligibility criteria most valuable---the inclusion of eligibility criteria no longer limits the data to the evaluation of supervised binary classification but opens its application to question-answering or language inference tasks.

We carefully examine the subset of SLRs produced by Cochrane, aiming to identify potential enhancements and extensions that would help mitigating the existing limitations of previous datasets. %
Every Cochrane SLR first registers and publishes the protocol containing the review title, abstract, search strategy and the eligibility criteria.
This information is all that human experts need to produce the final review, i.e., they first find the relevant studies and then conduct the meta-analysis of their results.
As described in Section~\ref{sec:task_formulation}, the screening process can be also modelled as a question-answering, where every publication is compared against the eligibility criteria in order to make the decision about the inclusion,\footnote{In the current approach, we consider only binary relevance (included versus excluded). However, in the real life, more categories can be defined reviewers (e.g. a study can be assigned as a background publication or meta-analysis).}
similar to the clinical decision support task of matching clinical trials to patients~\cite{roberts2017overview,roberts2021overview}.

To expand \datasetname, we searched the Cochrane Library\footnote{\url{https://www.cochranelibrary.com/cdsr/reviews}} for all SLRs from the meta-dataset based on the Cochrane review ID and take their latest open-access version.
We extract available information about the review: review title and abstract, eligibility criteria, search strategy and references.
Cochrane reports a list of included and excluded publications at the full text screening stage (this can be treated as approximately all included publications during the title and abstract screening stage).
Moreover, all excluded publications have a reason for exclusion selected by a reviewer.
As the original relevance judgements were limited to publications from the PubMed database, we assign PubMed IDs to these publications. %
We also define appropriate BigBio data loaders so the task can be seen as question-answering or textual pairs classification task.

\begin{table}[ht]
\centering
\caption{Details of the \datasetname~expanded meta-dataset. Column `\#docs' refers to the total number of documents included in all SLRS within the dataset, `\#included' mentions number of included documents on the title and abstract screening stage and `Avg. \%included' the percentage of included  publications averaged from all reviews.}
\label{tab:csmed_test_details}
\resizebox{0.99\textwidth}{!}{%
\begin{tabular}{@{}lrrrrrr@{}}
\toprule
Dataset name         & 
\#reviews & 
\#docs & 
\#included & 
\begin{tabular}[c]{@{}c@{}}Avg.\\\#docs\end{tabular} &
\begin{tabular}[c]{@{}c@{}}Avg. \%\\included\end{tabular} &
\begin{tabular}[c]{@{}c@{}}Avg. \#words\\in document\end{tabular} \\ \midrule
\textsc{\datasetname-train-basic}    & 30                    & 128,438             & 7,958  & 4,281             &  9.6\%        & 229    \\
\textsc{\datasetname-train-cochrane} & 195                   & 372,422             & 7,589     & 1,910       & 21.9\%      & 180   \\
\textsc{\datasetname-dev-cochrane} & 100                   & 229,376             & 4,365      & 2,294       & 20.8\%       & 201   \\ \midrule
\textsc{\datasetname-all} & 325                   & 730,236             & 19,912    & 2,247    & 20.5\%            & 195   \\
\bottomrule
\end{tabular}
}
\end{table}

Details of the new expanded \datasetname~are provided in Table \ref{tab:csmed_test_details}.
We were not able to find suitable data for all SLRs, hence the expanded \datasetname~is smaller than the original meta-dataset.
In total, the new expanded dataset consists of 295 unique Cochrane SLRs and 30 non-Cochrane SLRs.
The entire set of basic SLRs is designated for training. 
From the Cochrane reviews, we randomly selected 195 to the training split and the remaining 100 to the development split.
We abstain from designating a test split because \datasetname~aggregates existing datasets. 
Given the constraints of these datasets, creating a new, unbiased test collection is recommended.

\subsection{Baseline experiments}

We evaluate two models in a zero-shot setting: the traditional BM25 ranking function and the MiniLM-L6-v2\footnote{\url{https://huggingface.co/sentence-transformers/all-MiniLM-L6-v2}} Transformer-based model implemented in the retriv Python package~\cite{Bassani_retriv_A_Python_2023}. 
Predictions are run on the \textsc{\datasetname-dev-cochrane} split, and we test three different input representations using the following sections from the SLR protocol: (1) title, (2) abstract, and (3) eligibility criteria section.
We evaluate models using $nDCG@10$, $MAP$, Recall at rank $k$ with $k$ in \{10, 50, 100\} ($R@k$).  
Additionally, we compute three measures specifically designed for the task of CS: True Negative Rate at 95\% Recall ($TNR@95\%$)~\cite{KUSA2023200193,kusa2023vombat}, normalised Precision at 95\% Recall ($nP@95\%$)~\cite{kusa2023vombat}, and average position at which the last relevant item is found~\cite{Kanoulas2017CLEFOverview,Kanoulas2018CLEFOverview,Kanoulas2019CLEF2T}, calculated as a percentage of the dataset size, where a lower value indicates better performance ($Last~Rel$).

\begin{table*}[ht]
\centering
\caption{
Baseline results on \textsc{\datasetname-dev-cochrane} dataset.
\textbf{Bold} values indicate best score.
} 
\label{tab:csmed-cochrane-dev-results}
\resizebox{0.87\textwidth}{!}{
\begin{tabular}{ll|ccccccccc}
\toprule
\textbf{Model} & \textbf{Representation} & \textbf{TNR@95\%} & \textbf{nP@95\%} & \textbf{Last Rel} & \textbf{NDCG@10} & \textbf{MAP} & \textbf{R@10} & \textbf{R@50} & \textbf{R@100} \\
\midrule
\multirow{3}{*}{BM25} & Title & 0.469 & 0.142 & 72.2 & 0.438 & 0.388 & 0.349 & 0.623 & 0.704 \\
 & Abstract & 0.474 & 0.170 & \textbf{63.6} & 0.503 & 0.453 & 0.379 & 0.657 & 0.757 \\
 & Criteria & 0.404 & 0.122 & 84.9 & 0.286 & 0.258 & 0.224 & 0.418 & 0.510 \\
\midrule
\multirow{3}{*}{MiniLM} & Title & 0.472 & 0.217 & 68.1 & 0.470 & 0.414 & 0.379 & 0.673 & 0.763 \\
 & Abstract & \textbf{0.492} & \textbf{0.240} & 65.5 & \textbf{0.517} & \textbf{0.451} & \textbf{0.398} & \textbf{0.682} & \textbf{0.766} \\
 & Criteria & 0.347 & 0.144 & 75.4 & 0.299 & 0.299 & 0.233 & 0.492 & 0.588 \\
\bottomrule
\end{tabular}
}
\end{table*}

Table~\ref{tab:csmed-cochrane-dev-results} presents the baseline results.
Both for the BM25 and MiniLM models, using the systematic review abstract text as a query representation achieved the highest performance in all metrics compared to using the SLR title and criteria sections.
The MiniLM models consistently outperform their BM25 variant for all query representations on all evaluation measures except Last Relevant found.
The best-performing model, MiniLM using SLR abstracts, achieves $TNR@95\%$ equal to almost 0.5, meaning that this model can remove, on average, almost half of the true negatives when achieving a recall of 95\%.
Interestingly, despite the significance attributed to the eligibility criteria section by reviewers during paper screening, these models seemed challenged in leveraging the criteria information. 
It should be noted, however, that this section is typically more relevant to full-text screening rather than title and abstract. 
Exploring more advanced language models might reveal the potential for using this underutilised information.

\section{\ftdatasetname: Full text classification dataset} \label{sec:full_text_dataset}

In this section, we introduce \ftdatasetname~full text screening dataset and present baseline experiments.%

\subsection{Dataset creation}

LLM advancements have enabled processing long text snippets~\cite{beltagy2020longformer,zaheer2020big,mohtashami2023landmark,guo2021longt5}. Commercial tools now support inputs of up to 32k~\cite{openai2023gpt4} or even up to 100k tokens~\cite{anthropic}.
We propose \ftdatasetname, the full text screening dataset to enable research associated with the comprehensive understanding of very long documents, and evaluate such capabilities. We first gather full text versions of publications from \datasetname~SLRs, and then create the appropriate setting with canonical splits.

We use SemanticScholar and CORE~\cite{Knoth2023-zi,oro35755}
APIs to find URLs to open-access full text documents. 
This process successfully finds URLs to, on average, 27\% of all included and excluded publications from SLRs.
After downloading full text PDFs, we use GROBID~\cite{GROBID} to parse the content of these documents into an \textsc{xml} format.

We establish canonical splits considering the timestamps, such that the newest reviews belong to the test set.
Specifically, we select 31 Cochrane reviews published in the last year (between 01/06/2022 and 31/05/2023) to create a test set, another 60 reviews (mentioned in \citet{NUSSBAUMERSTREIT20181}) for the development set, and 176 reviews (listed by \citet{Scells2017}) as the training set.
Filtering out reviews with no associated available full text publications %
results in 148/36/29 reviews in train/dev/test splits.

Details of \ftdatasetname~are presented in Table~\ref{tab:csmed_ft_details}.
We also release a subset of 50 randomly selected documents from the test set as \textsc{\ftdatasetname-test-small}. At the moment of writing this publication, creating a prompt for LLMs with an input of few thousands tokens is feasible albeit costly,\footnote{According to the OpenAI model \href{https://openai.com/pricing}{pricing} on 22/09/2023, screening 500 full text documents with the GPT-4-32k model would cost more than 400 USD.} See Appendix \ref{app:csmed_ft_details} for further details on the creation of \ftdatasetname.

\begin{table}[ht]
\centering
\caption{Details of the \ftdatasetname~dataset. Column `\#docs' refers to the total number of documents included in the dataset and `\#included' mentions number of included documents on the full text step.  \textsc{\ftdatasetname-test-small} is a subset of \textsc{\ftdatasetname-test}.}
\label{tab:csmed_ft_details}
\resizebox{0.85\textwidth}{!}{%
\begin{tabular}{@{}lrrrrrr@{}}
\toprule
Dataset name         & 
\#reviews & 
\#docs. & 
\#included & 
\begin{tabular}[c]{@{}c@{}}\% included\end{tabular} &
\begin{tabular}[c]{@{}c@{}}Avg. \#words\\in document\end{tabular} &
\begin{tabular}[c]{@{}c@{}}Avg. \#words\\in review\end{tabular}\\ \midrule
\textsc{\ftdatasetname-train}       & 148                    & 2,053          & 904         & 44.0\%   & 4,535  & 1,493     \\
\textsc{\ftdatasetname-dev}         & 36                    & 644          & 202         & 31.4\%   & 4,419 & 1,402        \\
\textsc{\ftdatasetname-test}        & 29                    & 636          & 278         & 43.7\%   &  4,957  & 2,318     \\
\textsc{\ftdatasetname-test-small}      & 16                    & 50           & 22            & 44.0\%  &  5,042   & 2,354             \\ \bottomrule
\end{tabular}
}
\end{table}

\ftdatasetname~could be a proxy for a very long document natural language inference (NLI) task.
Popular NLI datasets (SciTail~\cite{khot2018scitail}, McTest~\cite{richardson-etal-2013-mctest} or DocNLI~\cite{yin-etal-2021-docnli}) contain both hypotheses and premises of an average length considerably shorter than 1,000 words; whereas
 in \ftdatasetname, the premise (review protocol) has an average length of more than 1,000 words, and the hypothesis (publication) contains more than 4,000 words.

\subsection{Experiment} \label{sec:experiments}

We present how \ftdatasetname~can be used to evaluate LLMs capabilities in making eligibility decisions on very long documents.
We run experiments both on fine-tuning of Transformer models and zero-shot prompting of GPT models.

\paragraph{Model selection}
As the combined input size of systematic review and publication can be big (9,246 mean number of tokens on a training split measured with a GPT-4 tokeniser), we only select models that allow inputs of at least 4k tokens context.
We fine-tune the open-domain Longformer and BigBird, and domain-specific models pre-trained on clinical data: Clinical-BigBird and ClinicalLongformer.
For zero-shot evaluation, we select GPT-3.5-turbo-0301, GPT-4-8k and GPT-3.5-turbo-16k accessed via OpenAI API.
GPT-4-8k and GPT-3.5-turbo-16k are the only models capable of handling more than 4k-input tokens, with context window size of 8k and 16k tokens respectively.

\paragraph{Preprocessing and evaluation}
For all models, we concatenate the screening protocol with each publication; we truncate the review description text to half of the available context window (2,000 tokens for 4k models, 4,000 tokens for 8k model and 8,000 tokens for 16k model) and complete the input with a publication.

For GPT models, if a whole publication text does not fit the context window, we run multiple predictions with a sliding window and aggregate the results.
In the case of GPT-3.5-turbo-16k model, only for 4 out of 50 documents the model was unable to process the full text of combined review and publication inside one prompt.

We fine-tune the Transformer models on \textsc{\ftdatasetname-train} for four epochs and run evaluation on \textsc{\ftdatasetname-dev}.
Due to the budget limitation, for the GPT-4-8k model, we run the evaluation only on \textsc{\ftdatasetname-test-small} (see Appendix~\ref{app:experiments} for further details on experimental settings). 
Finally, we evaluate the models using macro-averaged Precision, Recall and F1-score measures.

\paragraph{Results}
Results of the full text experiment are summarised in Table~\ref{tab:full-text-results}.
On \textsc{\ftdatasetname-test-small}, GPT-4-8k strongly outperforms other models. However, this difference is not statistically significant.
The GPT-3.5-turbo-16k achieves the highest Precision; this improvement can be attributed to the model's expanded context window and the limitations other GPT-based models have with our simple aggregation rules.
However, this might also be caused by overfitting towards the positive class, as this model includes almost twice as many publications as other models.
On \textsc{\ftdatasetname-test} set, Clinical-BigBird, significantly outperforms zero-shot GPT-3.5 model and pre-trained models based on the LongFormer architecture.

Interestingly, both BigBird-based models outperform their counterparts using the Longformer architecture.
The typical overall tendency to domain-pre-trained models achieving higher scores over their open-domain counterparts is also preserved.
We believe that fine-tuning the Transformer models first on larger NLI/QA corpora could help improve the results.

\begin{table}[ht]
\centering
\caption{Results of the full text screening experiment averaged over documents. 
The statistical significance was assessed with a McNemar’s t-test (p < 0.05) with Bonferroni correction for multiple testing.
\emph{Clinical-BigBird} on the \textsc{\ftdatasetname-test} split showed statistically significant improvements compared to the \emph{stratified random} baseline, \emph{Longformer}, \emph{Clinical-Longformer}, and \emph{GPT-3.5-turbo-16k}, indicated by $^\dag$.
Stratified baseline is averaged from 100 different random seeds. 
`\% incl.' describes the percentage of documents predicted as relevant by models.
}
\label{tab:full-text-results}
\resizebox{0.93\textwidth}{!}{%
\begin{tabular}{@{}l|llll|llll@{}}
\toprule
            & \multicolumn{4}{l|}{\textsc{\ftdatasetname-test-small}}      & \multicolumn{4}{l}{\textsc{\ftdatasetname-test}}        \\ \midrule
            & \% incl. & Precision & Recall & F1-score & \% incl. & Precision & Recall & F1-score  \\ \midrule
\textsc{oracle}  & 44\% & ---     & ---  & ---   & 43.7\%  & ---     & ---  & ---       \\
stratified random  & 50\% & 0.497     & 0.498  & 0.495 & ---    & 0.499     & 0.499  & 0.498       \\
`include all' & 100\% & 0.220     & 0.500  & 0.306  & 100\%   & 0.219     & 0.500  & 0.304       \\ \midrule
Longformer~\cite{beltagy2020longformer}           & 40\%  &    0.467       &  0.468       &  0.466     & 40.4\%     &  0.398        & 0.400          &  0.398        \\
BigBird-roberta-base~\cite{zaheer2020big}        &42\%     &    0.572       &  0.571       &   0.572   & 45.1\%       &  0.575        & 0.575          &  0.575        \\
Clinical-Longformer~\cite{li2022clinical}        & 36\%     &  0.547        &  0.544        &   0.542  & 35.1\%       & 0.436       &  0.441        &  0.435         \\ 
Clinical-BigBird~\cite{li2022clinical}         & 36\%    &   0.590        &  0.584        &  0.583    & 32.8\%      &   \textbf{0.623}$^\dag$     & \textbf{0.611}$^\dag$         & \textbf{0.609}$^\dag$         \\ \midrule
GPT-3.5-turbo-0301       & 54\%      & 0.585          &  0.586      &  0.580      & ---        & ---          &  ---      &  ---                   \\
GPT-4-8k-0314         & 58\%    &   0.674        &   \textbf{0.672}     &   \textbf{0.660}    & ---        &  ---         &  ---      &  ---                  \\ 
GPT-3.5-turbo-16k       & 80\%      &   \textbf{0.712}        &   0.638     &   0.576     & 75.9\%       &  0.538         &  0.528      &  0.475                  \\ \bottomrule
\end{tabular}
}

\end{table}

\section{Discussion} \label{sec:discussion}

In this paper, we have addressed the challenge of standardised evaluation in CS automation. 
By revisiting existing screening datasets, we evaluated their suitability as benchmarks in the context of modern ML methods. 
Our analysis revealed limitations such as small size and data leakage issues.

To overcome these challenges, we introduced \datasetname, a meta-dataset consolidating nine publicly released collections, providing programmatic access to \reviewscount~SLRs.
\datasetname~serves as a comprehensive resource for training and evaluating automated citation screening models and can be used for tasks that involve textual pairs classification, question answering and NLI. 
Additionally, we included a new dataset within \datasetname~for evaluating full text publication classification and conducted initial experiments showing that there is a room for improvement in understanding long contexts.

The focus of \datasetname~on providing unified access over a number of diverse citation screening datasets has many benefits.
First, the evaluation code can be re-used, making sure that the evaluation is handled properly.
Secondly, integration with the BigBio framework enables quick prototyping of prompts.
We also improve the documentation for existing datasets and provide a comprehensive data card for \ftdatasetname.
Our extended version of \datasetname~is also deduplicated.
Finally, it is a step towards providing a multi-domain SLR dataset and bridging the gap between IR and NLP research in the domain of screening automation, enabling direct comparisons of the methods.

\paragraph{Limitations and future work} \label{sec:limitations}
While we attempted to extract the data protocols as accurately as possible, extraction of data was not possible for all previous reviews.
This was primarily due to the changing standards in Cochrane reviews throughout the years.
In future work, ideally, direct access to Cochrane metadata would be needed to make sure that all information is covered.
Even though the PubMed publications most probably will not change, what can change is the API and scripts necessary to download the data.
There exists also the possibility that one of the sources will introduce a restriction on using their data for training and evaluation of machine learning models. %
We tried to further mitigate this potential issue by selecting open-access SLRs produced by Cochrane.
Finally, we acknowledge that using machine assistance for citation selection can raise concerns about research quality, emphasising the vital role of human oversight throughout the process.

Future work will focus on further improving data quality, connecting the output reviews from screening tools like \emph{CRUISE-Screening}~\cite{kusa2023cruise}, adding datasets covering other domains and different SLR tasks and designing a dataset for a prospective evaluation of review automation which could ensure no data leakage~\cite{cohen2010prospective}.
For the prospective dataset, predictions could be made as soon as the protocol is published, and the gold standard data becomes available when the review is eventually published, albeit with the drawback of a potentially long waiting time for review publication.

\section{Conclusion}

Our paper introduces \datasetname, a meta-dataset that addresses the lack of standardisation in SLR automation. 
By consolidating datasets and providing a unified access point, \datasetname~facilitates the development and evaluation of automated citation screening and full text classification models. 
We believe it has the potential to advance the field and lead to more robust automated SLR systems.
We envision \datasetname~as a living, evolving collection, and we invite researchers to contribute to expanding it with SLR datasets from other domains.

\begin{ack}
This work was supported by the EU Horizon 2020 ITN/ETN on Domain Specific Systems for Information Extraction and Retrieval -- DoSSIER (H2020-EU.1.3.1., ID: 860721).
\end{ack}

{
\small
\bibliography{anthology,references,suppl_final}
\bibliographystyle{plainnat}
}

\appendix

\newpage
\section*{Appendix overview}

This section provides an overview of the supplementary materials required by NeurIPS for our submission.

Appendix~\ref{app:datasets} offers an extended literature review encompassing citation screening datasets, evaluation measures used, and dataset coverage for other systematic literature review (SLR) steps.
Appendix~\ref{app:visualisations} presents detailed descriptions of the visualisations we have created.
Appendix~\ref{app:ec2s_datasheet} provides documentation for the \datasetname~meta-dataset, including the datasheet. 
For the \ftdatasetname~dataset, please refer to Appendix~\ref{app:csmed_ft_details} for its detailed documentation.
In Appendix~\ref{app:dataset_overlap}, we delve into the specifics of dataset overlap, while Appendix~\ref{app:experiments} contains the experimental details.

All data loaders and data preprocessing scripts for \datasetname~are available under the following URL: \url{https://github.com/WojciechKusa/systematic-review-datasets}.
\ftdatasetname~can also be accessed under the following URL: \url{https://github.com/WojciechKusa/systematic-review-datasets/raw/main/data/CSMeD/CSMeD-FT.zip}

\clearpage
\section{Detailed literature review of datasets} \label{app:datasets}

We base our literature review on three recent surveys, which we extend to cover the results until May 2023:
\begin{itemize}[leftmargin=14pt]
    \item Systematic review conducted by~\citet{OMara-Eves2015} in 2015.
    \item Update to the review above, completed by~\citet{Norman2020} in 2020.
    \item Systematic review conducted by~\citet{VanDinter2021b} in 2021.
\end{itemize}

\subsection{Citation screening datasets}

We searched Google Scholar and Semantic Scholar for publications introducing new datasets for the citation screening task.
We then searched for the forward citations of the original publication to find usages of the datasets.
From our list, we excluded private datasets used in only one publication.
We found 12 datasets fulfilling the criteria.\footnote{Between the submission of the main paper and the supplementary materials, one more new citation screening dataset with 10 SLRs was released on 5 June 2023~\citep{BADAMI2023102231}.}
Table~\ref{tab:sysrev_datasets_comparison} presents a summary of these datasets.

\begin{table}[ht]
\centering
\caption{Systematic literature review datasets with their characteristics, sorted by the publication year. We included all publicly available datasets and private datasets which were used in more than one publication.}
\label{tab:sysrev_datasets_comparison}
\resizebox{\textwidth}{!}{%
\begin{tabular}{@{}rrrcclc@{}}
\toprule
    & \begin{tabular}[c]{@{}c@{}}Publication\end{tabular} & \# reviews & Domain         & \begin{tabular}[c]{@{}c@{}}Data\\ URL\end{tabular}                                                       & \begin{tabular}[c]{@{}c@{}}Publicly\\available\end{tabular} & \begin{tabular}[c]{@{}c@{}}In\\ \datasetname\end{tabular} \\ \midrule
(1)  & \citet{Cohen2006}, 2006                             & 15         & Drug           & \href{https://dmice.ohsu.edu/cohenaa/systematic-drug-class-review-data.html}{URL}                        & $\checkmark$                                                & $\checkmark$                                              \\
(2)  & \citet{wallace2010semi}, 2010                       & 3          & Clinical       & \href{https://github.com/bwallace/citation-screening}{URL}                                               & $\checkmark$                                                & $\checkmark$                                              \\
(3) & \citet{Miwa2014ReducingScreening}, 2014             & 4          & Social science & ---                                                                                                      & ---                                                         & ---                                                       \\
(4)  & \citet{Howard2016a}, 2016                           & 5          & Mixed          & \href{https://systematicreviewsjournal.biomedcentral.com/articles/10.1186/s13643-016-0263-z\#Sec30}{URL} & $\checkmark$                                                & $\checkmark$                                              \\
(5)  & \citet{Scells2017}, 2017                            & 93         & Clinical       & \href{https://github.com/ielab/SIGIR2017-SysRev-Collection}{URL}                                         & $\checkmark$                                                & $\checkmark$                                              \\
(6)  & \citet{Kanoulas2017CLEFOverview}, 2017              & 50         & DTA            & \href{https://github.com/CLEF-TAR/tar/tree/master/2017-TAR}{URL}                                         & $\checkmark$                                                & $\checkmark$                                              \\
(7)  & \citet{Kanoulas2018CLEFOverview}, 2018              & 30         & DTA            & \href{https://github.com/CLEF-TAR/tar/tree/master/2018-TAR}{URL}                                         & $\checkmark$                                                & $\checkmark$                                              \\
(8)  & \citet{Kanoulas2019CLEF2T}, 2019                    & 49         & Mixed          & \href{https://github.com/CLEF-TAR/tar/tree/master/2019-TAR}{URL}                                         & $\checkmark$                                                & $\checkmark$                                              \\
(9)  & \citet{Alharbi2019}, 2019                           & 25         & Clinical       & \href{https://github.com/Amal-Alharbi/Systematic_Reviews_Update}{URL}                                    & $\checkmark$                                                & $\checkmark$                                              \\
(10) & \citet{parmar2021automation}, 2021                  & 6          & Biomedical     & ---                                                                                                      & ---                                                         & ---                                                       \\
(11)  & \citet{wang2022little}, 2022                        & 40         & Clinical       & \href{https://github.com/ielab/sysrev-seed-collection}{URL}                                              & $\checkmark$                                                & ---                                                       \\
(12) & \citet{hannousse2022semi}, 2022                     & 7          & Comp. Science  & \href{https://github.com/hannousse/Semantic-Scholar-Evaluation}{URL}                                     & $\checkmark$                                                & $\checkmark$                                              \\ \bottomrule
\end{tabular}
}
\end{table}

A dataset created by \citet{Cohen2006} containing 15 SLRs is the first and, up until today, one of the most commonly used to evaluate the effectiveness of machine learning models.
Since then, more datasets have been introduced, and starting in 2016, a new dataset was released almost every year.
All these datasets differ in the total number of reviews, subdomain, average review size, and percentage of included studies. 
However, the overall tendency shows a very high-class imbalance towards the negative class (i.e., irrelevant publications).
Datasets introduced by \citet{parmar2021automation} and \citet{Miwa2014ReducingScreening} are not publicly available, yet they were used in two and three research papers, respectively, so we included them in our comparison.

Until 2017 all of the datasets contained only the citation list with eligibility decisions \citep{Norman2020}.
More recently, datasets started to include titles of SLRs and search queries used for finding publications.
Additional metadata is limited to search queries~\cite{Scells2017}, review protocols (three datasets released as a part of the CLEF TAR shared-task by~\citet{Kanoulas2017CLEFOverview,Kanoulas2018CLEFOverview,Kanoulas2019CLEF2T}), review updates~\cite{Alharbi2019} and seed studies \cite{wang2022little}.
However, none of the datasets includes the eligibility criteria, the most critical section of SLR text used by manual annotators when assessing the relevance of publications.
They also do not contain the information about why a particular paper was excluded from the review.
Without this data, the automated citation screening problem cannot be tackled in any other way than a binary decision.
This is not the case in real life, as a typical SLR contains at least several exclusion and inclusion criteria, and the decision about every paper can be presented as a multi-dimensional relevance problem.

So far, there has been little attention to review automation outside of the medical domain. 
The only available datasets are four social science reviews by \citet{Miwa2014ReducingScreening}, and seven computer science reviews by \citet{hannousse2022semi}.
Compared to the general interest and rate of production of SLRs in other domains, this overall underrepresentation of benchmark datasets could be improved.
We also found one dataset containing one large SLR of environmental policies~\citep{10.1145/3511808.3557600}, which has a different scope and format than other datasets, so we decided not to include it in \datasetname~yet.

Papers from the ML and NLP domains, very often evaluate their approaches on datasets introduced by~\citet{Cohen2006}, which is, at the moment of writing this review, 17 years old. 
On the other hand, IR focused papers present their evaluation on CLEF TAR task datasets.

In terms of evaluation of classification approaches, 
aside from Precision and Recall, metrics include variations of the harmonised mean between the two, i.e. F$_\beta$--score, $utility$, $U19$~\cite{wallace2010semi,Wallace2010ActiveScreening,wallace2011should}, sensitivity-maximising thresholds~\cite{dalal2013pilot}, and $AUC$~\cite{cohen2010prospective}. 
Work Saved over Sampling ($WSS$) was proposed as a custom metric for evaluating this task as it measures the amount of work saved when using machine learning models to screen irrelevant publications \cite{Cohen2006,Matwin2010AReviews,Kontonatsios2020,Kusa2022AutomationStudy}.
The True Negative Rate ($TNR$) was proposed as an alternative as it addresses some of the limitations of WSS regarding averaging scores from multiple datasets \cite{KUSA2023200193}.
The measures of normalised Precision at r\% recall (nPrecision@r\%) and normalised rectified TNR at r\% recall  (nReTNR@r\%) have also been introduced to focus on other important aspects of screening task: screening full texts and estimating users' time savings when compared to the random ranking, respectively~\cite{kusa2023vombat}.

Cost-based and economic-based metrics were also used, especially in the context of the query formulation task in the CLEF TAR shared task \cite{Kanoulas2019CLEF2T,Kanoulas2017CLEFOverview,Kanoulas2018CLEFOverview}, e.g., total cost (TC) or total cost with a weighted penalty (TCW).
The TREC Total Recall track~\cite{grossman2016trec} also used a cut-off based metric, $recall @ aR + b$, which is defined as the recall achieved when $aR + b$ documents have been identified, where $R$ is the number of relevant documents in the collection and $a$ and $b$ are parameters. 
When $a = 1$ and $ b = 0$, $recall @aR + b$  is equivalent to R-precision. 
Finally, there has been a proposal to shift away from measuring Recall and instead evaluate how accurately automated methods can replicate the original systematic review outcomes~\cite{kusa2023outcome}.

The practical relevance of evaluating CS with metrics like the area under the ROC curve (AUC)~\cite{lee2023pgb} has been called into question, as it may not align with the goals of the citation screening task. 
Given that the CS task is primarily focused on achieving high recall, using AUC as an evaluation metric can be misleading, as it may highlight model improvements at lower recall values~\cite{KUSA2023200193}.
Having a unified benchmarking approach would also help to resolve these problems.

Finally, we were interested in checking how recently each dataset was used, where that usage was published, and what kind of evaluation measures were applied to that data.
Table~\ref{tab:dataset_usage} presents the summary of our findings. 
We can see that to this date, most datasets were used in the past two years and simultaneously used by different publications.
There is also a disparity in used evaluation measures, yet the basic Precision, Recall and F1-score prevail.

\begin{table}[h]
    \centering
    \caption{Usage statistics of the SLR datasets, including the latest publication year, venue and evaluation measure. We report two usages in case there was a more recent pre-print published.}    \label{tab:dataset_usage}
    \resizebox{\textwidth}{!}{%
    \begin{tabular}{rrlll}
    \toprule
      
      & 
      \multicolumn{2}{l}{Release - last time used}& 
   \begin{tabular}[l]{@{}l@{}}Evaluation schema (latest)\end{tabular} & \begin{tabular}[l]{@{}c@{}}Venue (latest)\end{tabular} \\ \midrule
     (1) & \multicolumn{2}{l}{2006 - 2023~\cite{lee2023sr,KUSA2023200193}}& TNR~\cite{KUSA2023200193}, AUC~\cite{lee2023sr} & ECIR\\  
     (2) & \multicolumn{2}{l}{2010 - 2022~\cite{Kusa2022AutomationStudy}}& WSS, Precision@95\%~\cite{Kusa2022AutomationStudy} & ECIR\\  
     (3) & \multicolumn{2}{l}{2014 - 2016~\citep{Hashimoto2016TopicReviews}}& Yield, Burden, WSS~\citep{Hashimoto2016TopicReviews} & JBI\\  
     (4) & \multicolumn{2}{l}{2016 - 2022~\cite{Kusa2022AutomationStudy}, 2023~\cite{lee2023pgb}}& WSS, Precision@95\%~\cite{Kusa2022AutomationStudy}, AUC~\cite{lee2023pgb} & ECIR\\  
     (5) & \multicolumn{2}{l}{2017 - 2018~\cite{scells2018generating}}& Precision, Recall, WSS~\cite{scells2018generating} & SIGIR\\  
     (6) & \multicolumn{2}{l}{2017 - 2023~\citep{10.1145/3539597.3573025}}& Precision, F1, Recall~\citep{10.1145/3539597.3573025} & WSDM\\  
     (7) & \multicolumn{2}{l}{2018 - 2023~\citep{10.1145/3539597.3573025}}& Precision, F1, Recall~\citep{10.1145/3539597.3573025} & WSDM\\  
     (8) & \multicolumn{2}{l}{2019 - 2022~\citep{Wang2021}, 2023~\cite{lee2023pgb}}& MAP, Precision, nDCG~\citep{Wang2021}, AUC~\cite{lee2023pgb}  & ECIR\\  
     (9) & \multicolumn{2}{l}{2019 - 2020~\citep{10.1093/jamia/ocaa148}}& Recall, Precision~\citep{10.1093/jamia/ocaa148} & JAMIA\\  
     (10) & \multicolumn{2}{l}{2021 - 2022~\cite{parmar-etal-2022-boxbart}}& F1-Score~\cite{parmar-etal-2022-boxbart} & NAACL \\   
     (11) & \multicolumn{2}{l}{2022 - 2023~\cite{wang2023can}}& Precision, F1, F3, Recall~\cite{wang2023can} & SIGIR\\  
     (12) & \multicolumn{2}{l}{2022 - 2022~\cite{hannousse2022semi}}& Recall, Precision, Macro F1, Accuracy~\cite{hannousse2022semi} & MedPRAI\\  \bottomrule
    \end{tabular}
    }
\end{table}

\subsection{SLR datasets in biomedical benchmarks}

Systematic literature reviews consist of multiple steps, and depending on the granularity, previous studies enumerated between four and up to 15 tasks that might be included in the SLR process \cite{Tsafnat2014SystematicTechnologies}.
High-level tasks include steps of preparation, followed by the search and appraisal of primary studies and then synthesis and write-up of the evidence.
According to~\citet{VanDinter2021b}, citation screening (selection of primary studies) was the step for which most of the automation-related research was happening.
Among other steps, the tasks of query formulation, information extraction, risk of bias assessment, and, more recently, text summarisation were also introduced.

\citet{marshall2016robotreviewer} introduced a large dataset with Cochrane reviews for the task of assessing the risk of bias -- a procedure aiming at establishing the quality of input studies.
\citet{nye2018corpus} proposed a PICO (Population, Intervention, Comparison and Outcome) extraction dataset containing 5,000 annotated abstracts of biomedical publications.
In the query formulation, often the models evaluate their performance on the CLEF TAR 2017-2018 datasets~\cite{Kanoulas2017CLEFOverview,Kanoulas2018CLEFOverview}.
For the task of systematic review summarisation, a shared task was introduced~\cite{wang-etal-2022-overview} consisting of two datasets: \citep{Wallace2020GeneratingN,DeYoung2021MS2MS}.

In a comprehensive catalogue of medical artificial intelligence datasets and benchmarks by~\citet{Blagec2022BenchmarkProfessionals}, only three citation screening datasets are mentioned:~\citet{Cohen2006}, ~\citet{Wallace2010ActiveScreening}, and~\citet{Miwa2014ReducingScreening}.
Of these three datasets, only two are publicly available, and both are already implemented in \datasetname.
Additionally, another five private SLRs used in only one publication~\cite{singh2018improving} are mentioned.

There is poor coverage of SLR datasets among biomedical benchmarks, especially for the task of citation screening.
None of the existing benchmarks contains any publicly available citation screening dataset. 
Only the BoX~\cite{parmar-etal-2022-boxbart} benchmark uses five SLRs, but these datasets are private and cannot be obtained even through a DUA (Data Use Agreement).

From other SLR automation tasks, BigBio~\cite{fries2022bigbio} and BLURB~\cite{DBLP:journals/health/GuTCLULNGP22} benchmarks contain only one information extraction dataset by~\citet{nye2018corpus}.
BLUE~\cite{peng-etal-2019-transfer} and CBLUE~\cite{zhang-etal-2022-cblue} benchmarks do not contain any SLR-related task.
Therefore, there is a clear need to develop and include publicly available SLR datasets in biomedical benchmarks, particularly for citation screening tasks, to facilitate further research and progress in this field.

The latest advances in Large language models (LLMs) offer significant potential for aiding in SLR automation but simultaneously raise several concerns. 
A user study by \citet{yun2023appraising} mentions that SLR practitioners acknowledged the potential utility of LLMs in various tasks, such as generating the first draft of a review, writing plain language summaries, and extracting information from longer texts. 
On the other hand, domain experts have highlighted several crucial issues, including concerns about hallucinations, the untraceable origins of generated content, and the proliferation of bad-quality reviews.

\section{Visualisations}
\label{app:visualisations}

\begin{figure}[ht!]
\centering
\includegraphics[width=\linewidth]{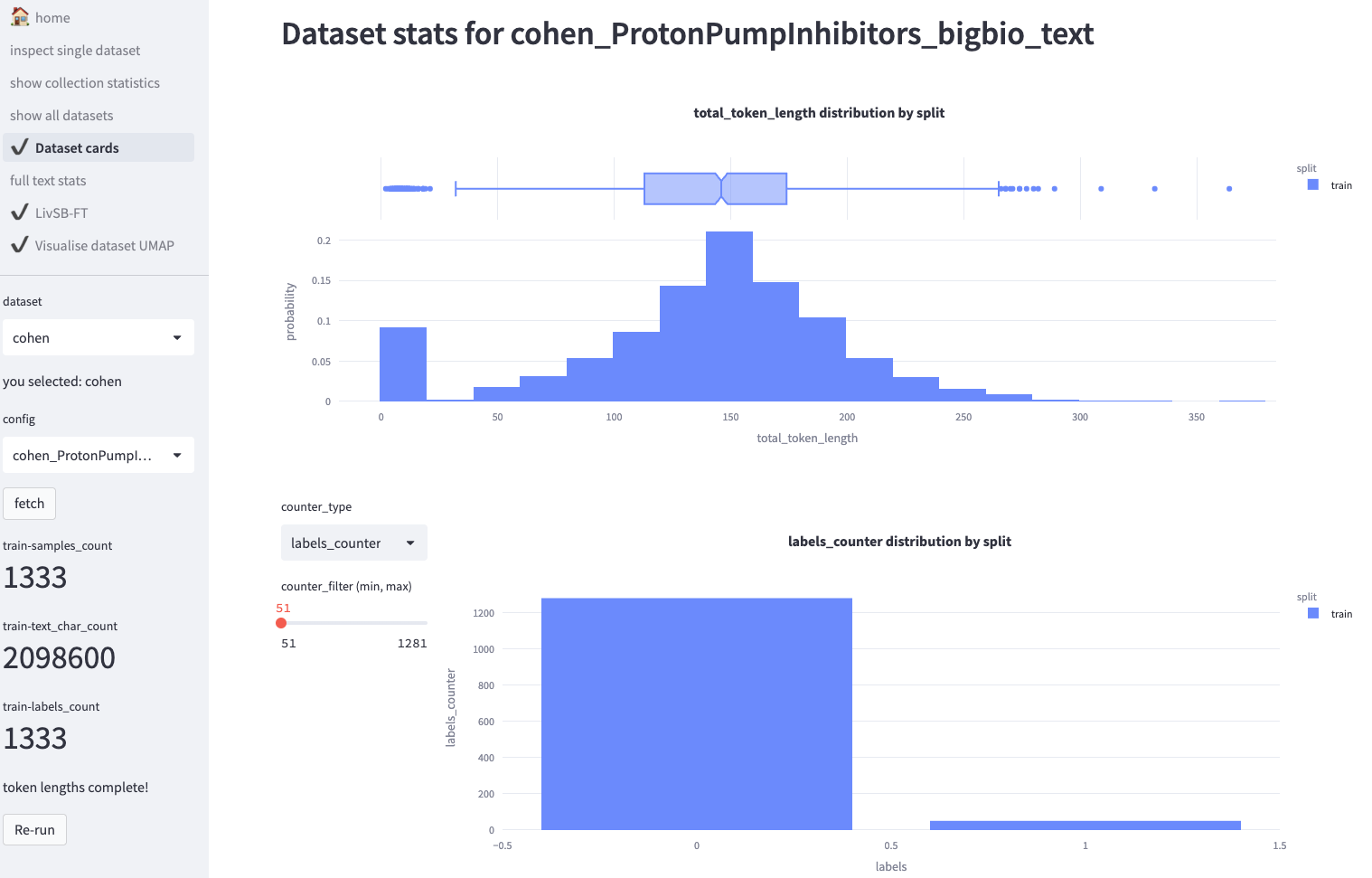}
\caption{\label{fig:vis_data_card}Example visualidation with statistics for a ``Proton Pump Inhibitors'' SLR dataset.}
\end{figure}

We leverage Streamlit\footnote{\url{https://streamlit.io}} to create interactive visualisations for our meta-dataset.
We present essential details for every dataset, such as the number of training samples, character and word counts, and labels and token lengths distribution across dataset splits (example on Figure~\ref{fig:vis_data_card}). 
We build upon the existing BigBio schemas and visualisations, extending them to incorporate citation screening-specific details.
We also build a dedicated page to explore \ftdatasetname~dataset containing full text documents.

We further focus on measuring the overlap between datasets. 
We check for the overlap on the level of systematic reviews based on the review's Cochrane ID.
This can help researchers understand potential biases, redundancy, or complementary aspects across various datasets.

\begin{figure}[ht!]
\centering
\includegraphics[width=\linewidth]{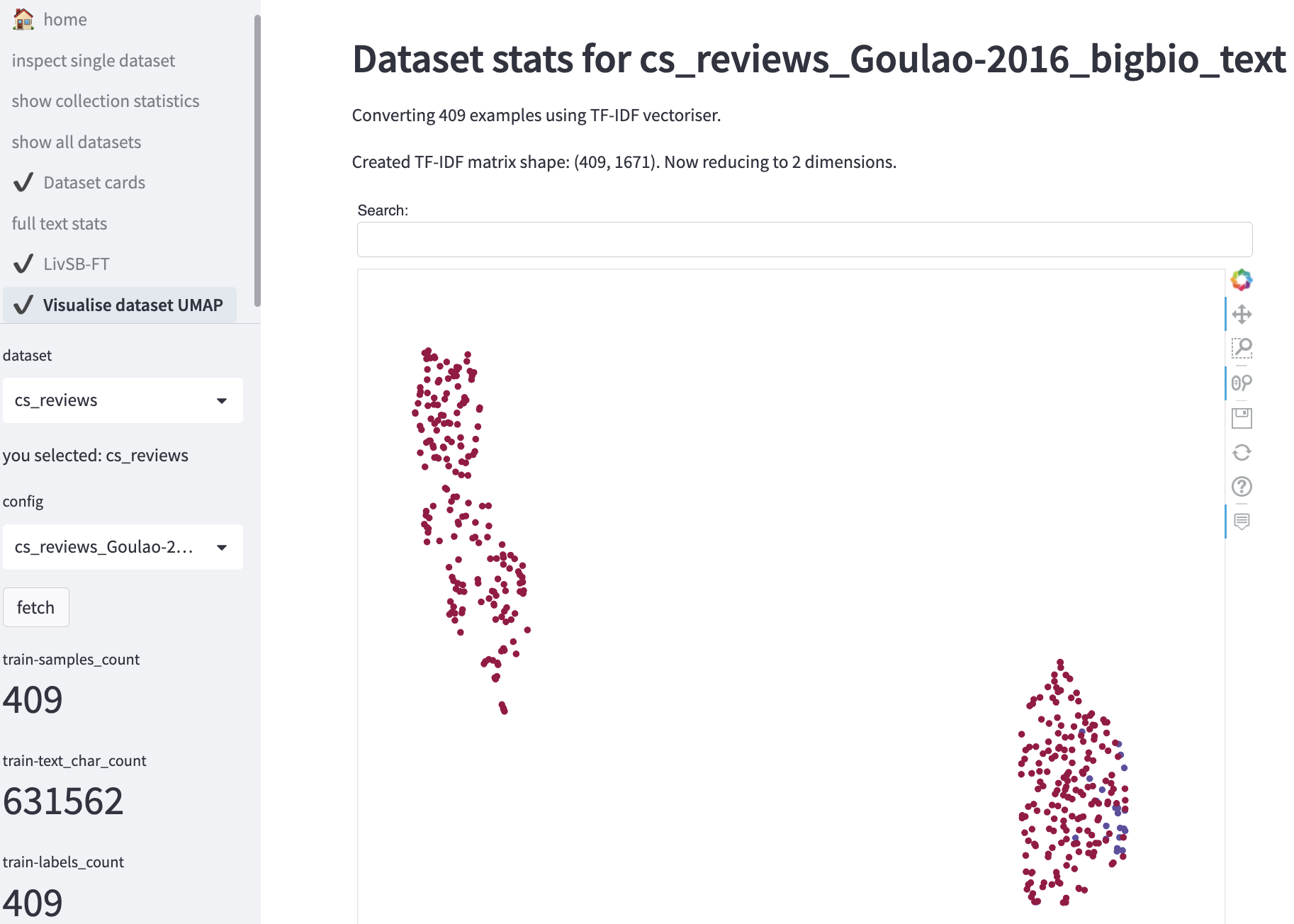}
\caption{\label{fig:vis_umap_cs}Example visualisations with TF-IDF and UMAP representation of documents for a ``CS-Goulao-2016'' SLR. 
Based on the plot, one can see that the retrieved documents are grouped in two clusters with all relevant publications belonging to one of them (bottom-right part of the plot).
This can be an indicator that any model will likely remove the other ``non-relevant'' cluster of documents and hence achieve good score in detecting true negatives.}
\end{figure}

We use a TF-IDF-based document vectoriser with UMAP~\cite{mcinnes2018umap-software} to plot two-dimensional representations of the datasets. 
This approach allows us to effectively capture and display the structural patterns and similarities within a single systematic literature review, aiding researchers in identifying clusters, outliers, and potential data correlations.
An example of UMAP clustering of publications is presented in Figure~\ref{fig:vis_umap_cs}.

A live demo of the visualisation interface is available under the following URL: \url{https://systematic-review-datasets.streamlit.app/}.
Some features require data preprocessing; they are unavailable in the demo but can be run locally using the code from the GitHub repository.

\section{\datasetname~data card}
\label{app:ec2s_datasheet}

\textbf{Dataset Description:} 
\datasetname~is a meta-dataset consisting of nine different citation screening datasets containing \reviewscount~systematic literature reviews (SLRs). 
Each systematic review consists of a list of publications that need to be classified as either \emph{relevant} or \emph{irrelevant}.
All datasets have data loader scripts providing programmatic access aligned with the BigBio framework and HuggingFace datasets library.
We preserve the original splits of the datasets.
We also generate data cards for every dataset which is part of the \datasetname.
\datasetname~allows for accessing independent datasets and single systematic reviews, which are part of each dataset.

\textsc{train-cochrane} split contains expanded metadata about systematic reviews such as systematic review title, abstract, eligibility criteria and search strategy.
\textsc{train-basic} is a set of SLRs for which such meta-data was unavailable and it is characterised by the systematic literature review title.
\textsc{train-cochrane} split is suitable for the tasks of question answering, natural language inference, and text pair classification.
\textsc{train-basic} is suitable only for the text classification task.

\textbf{Homepage:} \url{https://github.com/WojciechKusa/systematic-review-datasets}

\textbf{URL:} \url{https://github.com/WojciechKusa/systematic-review-datasets}

\textbf{Licensing:} CC BY 4.0

\textbf{Languages:}  English

\textbf{Tasks:} 
text classification ({\tt TXTCLASS}), 
question answering ({\tt QA}), 
natural language inference ({\tt NLI}),
text pairs classification ({\tt PAIRS}). 

\textbf{Schemas:} 
Text ({\tt TEXT}), 
Text pairs classification ({\tt PAIRS}). 
Question Answering ({\tt QA}), 
source ({\tt source}).

\textbf{Splits:} \textsc{train-basic}, \textsc{train-cochrane}, \emph{all}

\section{\ftdatasetname} 
\label{app:csmed_ft_details}

\ftdatasetname~is an extension of the \datasetname~meta-dataset that specifically focuses on the full text screening step in SLRs.
\ftdatasetname~is to the best of our knownledge, it is the first dataset explicitly targeted at the screening of the full text of publication.
While previously researchers already used full text screening labels from other datasets to evaluate their models, the input to these models constituted only the titles and abstracts of publications~\cite{Howard2020}.

\subsection{Dataset construction details}

To construct \ftdatasetname, we collected various elements of SLRs from the Cochrane Library website, including the title, abstract and eligibility criteria sections of the SLR and SLRs' appendix and references.
The appendix contains a search strategy, while the references list papers categorised as: ``studies included in the review'',  ``studies excluded from the review'', and ``additional references''. 
We decided to focus solely on the ``included'' and ``excluded'' categories as there is no definitive way to determine the intended meaning when researchers added papers as additional references. 
However, in future work, we plan to explore the possibility of extending the dataset to encompass publications from the ``additional references'' category.

To obtain the full texts of references, we used the DOI (Digital Object Identifier) of each publication. 
While some references directly provided the DOI, for others, we initially attempted to match them to PubMed IDs and then extracted the DOIs from PubMed and Semantic Scholar.
To assign PubMed IDs to the publications parsed from the Cochrane website, we followed a four-step process:
\begin{itemize}[leftmargin=14pt]
    \item We check if the PubMed ID information is provided on the Cochrane references webpage.
    \item We conduct search in PubMed using \textsc{Entrez}\footnote{\url{https://www.ncbi.nlm.nih.gov/search/}} by searching for the same title and authors.
    \item We search for the PubMed ID in SemanticScholar using publication DOI from Cochrane references webpage.
    \item We search again in PubMed, this time with a relaxed requirement by searching for an exact match in the title only.
\end{itemize}

We then use the PubMed ID to resolve the DOI of the publication. 
We could match the DOI for more than 61\% of references.

\begin{figure}[ht!]
\centering
\includegraphics[width=\linewidth]{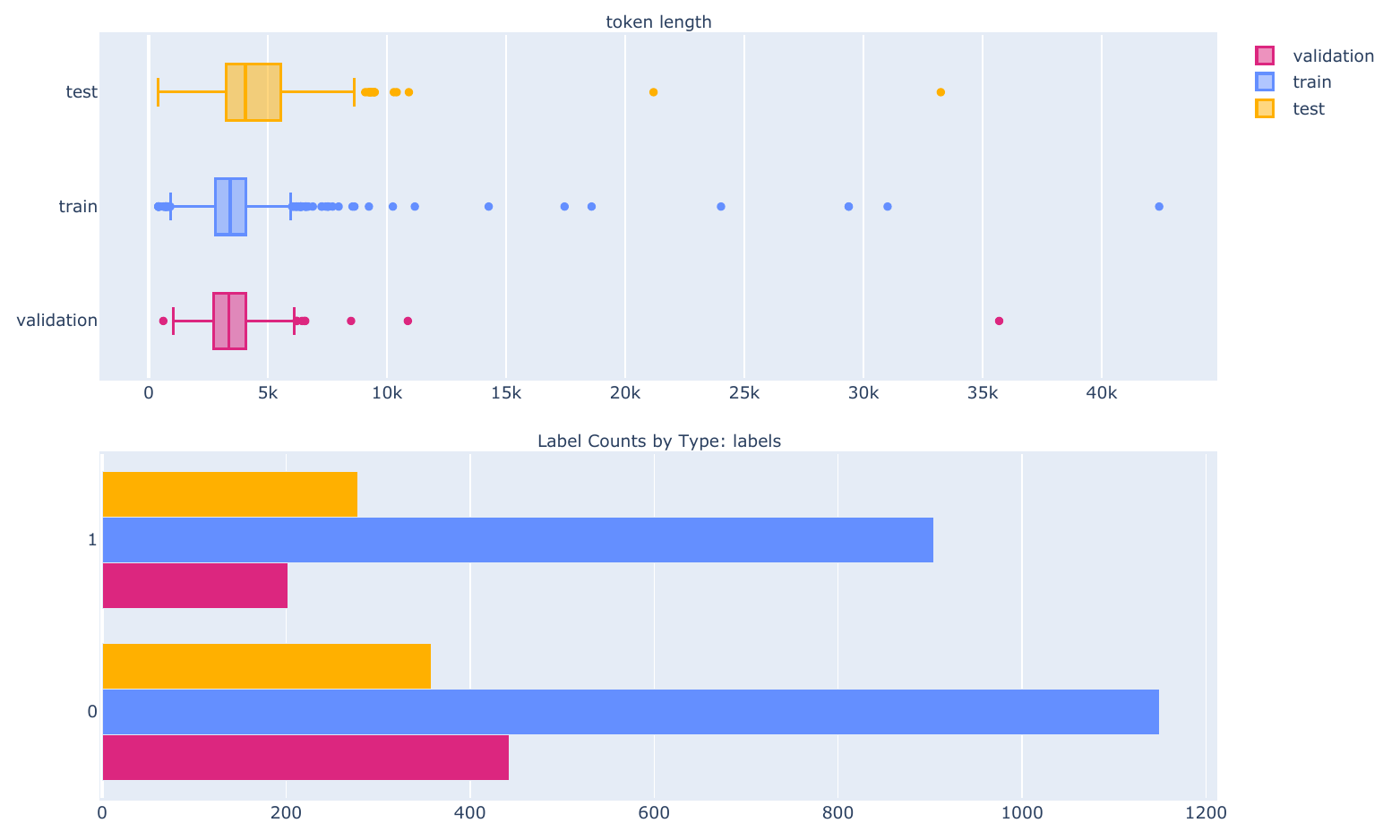}
\caption{\label{fig:csmed_ft}Token frequency distribution by split (top) and frequency of different kind of instances (bottom).}
\end{figure}

We adopted a time-wise construction approach for \ftdatasetname~canonical splits to ensure the integrity and avoid data contamination.
Therefore, we selected 29 SLRs not part of any previously released datasets to form our test set.
We used data from previous publications to construct a testing and development set: dataset used by \citet{NUSSBAUMERSTREIT20181} for the development set and dataset introduced by \citet{Scells2017} for training split.
It should be noted that newer SLRs tend to have more comprehensive metadata and more open-access full text publications available. 
This resulted in token length and label frequency differences across the dataset splits (Figure~\ref{fig:csmed_ft}).
Despite these variations, we decided to retain these splits as they present a more realistic and challenging scenario, closely reflecting real-life circumstances.

We have made the entire dataset construction procedure available in our repository, enabling transparency and reproducibility.

\subsection{\ftdatasetname~Data Card}

\textbf{Dataset Description} The dataset focused on the task of full text screening for systematic literature review creation.
It contains 3,333 systematic literature review and publication pairs with decisions if the publication was included in the systematic literature review.
Every excluded publication also contains a textual explanation of why it was excluded.
Systematic literature reviews are formatted in a \textsc{json} format, whereas publications are stored as \textsc{csv} files.
\textsc{\ftdatasetname-sample} is a subset of \textsc{\ftdatasetname-test} dataset.
We intend to store the dataset on the TU Wien Research Data repository,\footnote{\url{https://researchdata.tuwien.ac.at}} currently the dataset is available on the project GitHub repository.

\textbf{Homepage:} \url{https://github.com/WojciechKusa/systematic-review-datasets}

\textbf{URL:} \url{https://github.com/WojciechKusa/systematic-review-datasets/raw/main/data/CSMeD/CSMeD-FT.zip}

\textbf{Licensing:} CC BY 4.0

\textbf{Languages:} English

\textbf{Tasks:} text pairs classification, natural language entailment

\textbf{Schemas:} {\tt  TEXT}, {\tt PAIRS}, source.

\textbf{Splits:} \textsc{train}, \textsc{dev}, \textsc{test}, \textsc{sample}

\textbf{Dataset size (document pairs):} \textsc{train}: 2,053, \textsc{dev}: 644, \textsc{test}: 636, \textsc{sample}: 50

\textbf{Size of downloaded dataset files:} 33.5 MB

\textbf{Size of the generated dataset files:} 112.2 MB

\section{Examining dataset overlap}
\label{app:dataset_overlap}

We evaluate the overlap between datasets at the level of entire systematic reviews. 
This analysis aims to understand the potential duplication of information and data leakage across different datasets.

Table \ref{tab:dataset-overlap} presents the extent of overlap observed between the train and test splits of the datasets.
The TAR 2019 collection is most severely affected, with 3 SLRs duplicated in its train and test splits. 
SLRs released as part of the SIGIR 2017 collection~\cite{Scells2017} are also present among the test splits in CLEF TAR 2017 and 2019 collections.

\begin{table}[h]
    \centering
    \caption{List of overlapping Cochrane systematic literature reviews between datasets.}
    \label{tab:dataset-overlap}
\begin{tabular}{lll} 
\toprule Cochrane review ID & First collection & Other collections \\  \midrule 
CD011145 & sigir2017 (train) & tar2017 (test) \\ 
CD010633 & sigir2017 (train) & tar2017 (test), tar2018 (train), tar2019 (train) \\ 
CD010653 & sigir2017 (train) & tar2017 (test), tar2018 (train), tar2019 (train) \\ 
CD010542 & sigir2017 (train) & tar2017 (test), tar2018 (train), tar2019 (train) \\ 
CD009185 & sigir2017 (train) & tar2017 (test), tar2018 (train), tar2019 (train) \\ 
CD008081 & sigir2017 (train) & tar2017 (test), tar2018 (train), tar2019 (train) \\ 
CD002143 & sigir2017 (train) & sigir2017 (train) \\ 
CD001261 & sigir2017 (train) & tar2019 (test) \\ 
CD011571 & tar2019 (train) & tar2019 (test) \\ 
CD012164 & tar2019 (train) & tar2019 (test) \\ 
CD011686 & tar2019 (train) & tar2019 (test) \\ \bottomrule 
\end{tabular}

\end{table}

It is worth noting that we did not explicitly report the overlap between different CLEF TAR datasets~\cite{Kanoulas2017CLEFOverview,Kanoulas2018CLEFOverview,Kanoulas2019CLEF2T}. 
The owners of the dataset have already acknowledged that each new edition of the dataset includes SLRs from the previous editions as part of the training data. 
As the older datasets did not share metadata about the considered reviews (except for the very high-level title of the review (e.g. ADHD or COPD), we did not have access to the mapping to the published reviews.

\section{Experimental setup}
\label{app:experiments}

\subsection{Transformer model fine-tuning}

We select the following model checkpoints from HuggingFace Transformers library:

\begin{itemize}[leftmargin=14pt]
    \item Longformer-base -- \url{https://huggingface.co/allenai/longformer-base-4096}
    \item BigBird-roberta-base -- \url{https://huggingface.co/google/bigbird-roberta-base}
    \item Clinical-Longformer -- \url{https://huggingface.co/yikuan8/Clinical-Longformer}
    \item Clinical-BigBird -- \url{https://huggingface.co/yikuan8/Clinical-BigBird}
\end{itemize}

We want to decide whether a publication fulfils all inclusion criteria and none of the exclusion criteria to include it in the SLR.
Specifically, this means matching the eligibility criteria of SLR with the full text of the candidate publication. 
As input, the model receives the text of the review and publication and is asked to predict a binary category.
We concatenate the review title with the eligibility criteria section to create the review text.
For publications, we concatenate the title, abstract and the main text.

As available input text (review text + publication text) almost always exceeds the available context window of considered models (4,096 tokens), we use the following approach to allocate available space.
We use the {\tt TokenTextSplitter} method from the langchain library\footnote{\url{https://github.com/hwchase17/langchain}} with the gpt-3.5-turbo-0301 model to select the review text that would fit the context window.
We select at most half of the available context window, so in the context of all Transformer models, review text equals at most 2,048 tokens.
This action truncates some part of the eligibility criteria section, i.e. for 13\% of items in the trainset and 42\% in the test set (Table~\ref{tab:csmed-ft-tokeniser-stats}).
We fill the remaining input sequence with the publication text.

\begin{table}[h]
    \centering
    \caption{Statistics of a review text with respect to the fit within 2,048 tokens context window.}
    \label{tab:csmed-ft-tokeniser-stats}
\resizebox{\textwidth}{!}{%
\begin{tabular}{rccccc}
\toprule
 & \textsc{\ftdatasetname-train} & \textsc{\ftdatasetname-dev} & \textsc{\ftdatasetname-test} & \textsc{\ftdatasetname-sample} \\
\midrule
Avg \# splits & 1.13 & 1.24 & 1.83 & 1.74 \\
Median \# splits & 1 & 1 & 1 & 1 \\
Max \# splits & 2 & 2 & 4 & 4 \\
Min \# splits & 1 & 1 & 1 & 1 \\
More than 1 splits & 13\%  & 24\%  & 42\%  & 42\%  \\
\bottomrule
\end{tabular}
}
\end{table}

We run our experiments on a single server with 4 Nvidia RTX 3090 GPUs with 24GB of RAM each. 
We use a per-device batch size of 1 with eight gradient accumulation steps.
We test several learning rates with the best results for 1e-5, and we set weight decay to 0.01.
We use AdamW~\cite{loshchilov2018decoupled} with default values of $\beta_1 = 0.9$ and $\beta_2 = 0.999$. 
We evaluate models after each epoch on the validation set and select the model with the highest macro f1-score.

One training epoch took around 30 minutes both for BigBird and Longformer-based models.
For inference, Longformer architecture processed, on average, 2.9 samples per second, whereas BigBird models 2.65 samples per second.
Making predictions on the entire test split of 636 documents took less than 4 minutes for all models.

\subsection{Zero-shot language model evaluation}

Similarly, as for the fine-tuned classification models, we reserve at most half of the context window size for the systematic literature review description and fill the remaining tokens with the publication text.
We measure the text length using the OpenAI library tiktoken\footnote{\url{https://github.com/openai/tiktoken}}, which provides tokenisers for GPT-3.5 and GPT-4 models.
We use the \texttt{openai} python library version \texttt{0.27.7}, and use the default chat completion function parameters of temperature = 1 and top\_p = 1. 

We set our total budget to 50 USD and conduct the experiments only on the \textsc{\ftdatasetname-test-small} subset for GPT-4 model.
For the GPT-3.5-turbo-16k model, making predictions on all 636 examples of the \textsc{\ftdatasetname-test} split took 44 minutes.
However, this value was heavily influenced by the default OpenAI's rate limits of 180,000	tokens per minute for our organisation.
We use the following prompt template:

\paragraph{Input Template:} \
\begin{code}
Does the following scientific paper fulfill all eligibility criteria and \ 
should it be included in the systematic review? \ 
Answer `Included' or `Excluded'. \
Systematic review: "{{r.title}}" \n "{{r.criteria}}" \n\n \ 
Publication: "{{p.title}}" \n "{{p.abstract}}" \n "{{p.main_text}} \n\n \
Answer: 
\end{code}
\paragraph{Output Template:} \
\begin{code}
{{label}}
\end{code}
\paragraph{Answer Choices:} \
\begin{code}
Included ||| Excluded 
\end{code}

\end{document}